\documentclass[letterpaper]{article} 
\usepackage{aaai25}  
\usepackage{times}  
\usepackage{helvet}  
\usepackage{courier}  
\usepackage[hyphens]{url}  
\usepackage{graphicx} 
\usepackage{natbib}  
\usepackage{caption} 
\usepackage{algorithm}
\usepackage{algorithmic}
\usepackage{amsmath}
\usepackage{booktabs}

\usepackage[marginal]{footmisc}

\usepackage{newfloat}
\usepackage{listings}
\DeclareCaptionStyle{ruled}{labelfont=normalfont,labelsep=colon,strut=off} 
\floatstyle{ruled}
\newfloat{listing}{tb}{lst}{}
\floatname{listing}{Listing}
\title{MM-Tracker: Motion Mamba with Margin Loss for UAV-platform Multiple Object Tracking}
\author {
    Mufeng Yao\textsuperscript{\rm 1}\equalcontrib,
    Jinlong Peng\textsuperscript{\rm 2}\equalcontrib,
    Qingdong He\textsuperscript{\rm 2},
    Bo Peng\textsuperscript{\rm 3},
    Hao Chen\textsuperscript{\rm 1},
    Mingmin Chi\textsuperscript{\rm 1}\thanks{Corresponding authors.},
    Chao Liu\textsuperscript{\rm 1}\footnotemark[2],
    Jon Atli Benediktsson\textsuperscript{\rm 4}\footnotemark[2]
}
\affiliations {
    \textsuperscript{\rm 1}School of Computer Science,Shanghai Key Laboratory of Data Science, Fudan University\\
    \textsuperscript{\rm 2}Tencent Youtu Lab\\
    \textsuperscript{\rm 3}Shanghai Ocean University\\
    \textsuperscript{\rm 4}University of Iceland\\
    
    mfyao21@m.fudan.edu.cn, jeromepeng@tencent.com, yingcaihe@tencent.com,bpeng@shou.edu.cn,
    22210240115@m.fudan.edu.cn, mmchi@fudan.edu.cn,
    chaoliu20@.fudan.edu.cn,benedikt@hi.is
}

\usepackage{bibentry}

\begin{document}

\maketitle

\begin{abstract}
Multiple object tracking (MOT) from unmanned aerial vehicle (UAV) platforms requires efficient motion modeling. This is because UAV-MOT faces both local object motion and global camera motion. Motion blur also increases the difficulty of detecting large moving objects.
Previous UAV motion modeling approaches either focus only on local motion or ignore motion blurring effects, thus limiting their tracking performance and speed. To address these issues, we propose the Motion Mamba Module, which explores both local and global motion features through cross-correlation and bi-directional Mamba Modules for better motion modeling. To address the detection difficulties caused by motion blur, we also design motion margin loss to effectively improve the detection accuracy of motion blurred objects. Based on the Motion Mamba module and motion margin loss, our proposed MM-Tracker surpasses the state-of-the-art in two widely open-source UAV-MOT datasets. Code is available at: https://github.com/YaoMufeng/MMTracker
\end{abstract}

%
\section{Introduction}
Multiple object tracking (MOT) aims to identify objects in a given video and is widely used in computer vision~\cite{luo2021multiple,2021Analysis}, including applications in autonomous driving~\cite{geiger2013vision}, human-computer interaction~\cite{chandra2015eye}, and pedestrian tracking~\cite{peng2020chained,meinhardt2022trackformer}.
A typical MOT approach consists of two primary stages: detection and association~\cite{shuai2021siammot}. The detection stage identifies all objects in each frame, while the association stage matches objects across consecutive frames to establish complete trajectories for each object~\cite{liu2022multi}.
Recently, MOT in unmanned aerial vehicle (UAV) platforms has attracted extensive research interest.
Compared with conventional MOT~\cite{dendorfer2020mot20}, MOT in UAV view faces more challenges.
\begin{figure}[t]
  \centering
  \includegraphics[width=\linewidth]{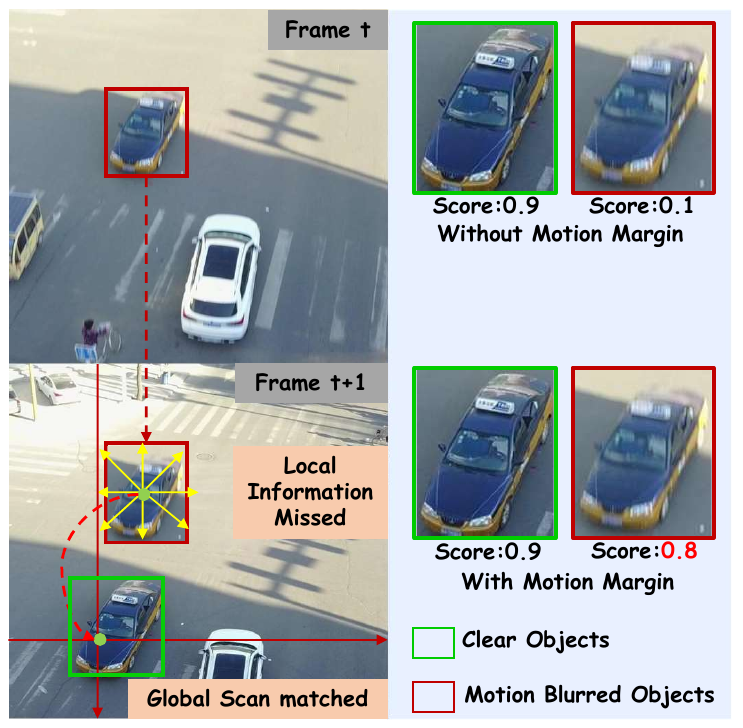}
  \vspace{-1.8em}
  \caption{Advantages of our MM-Tracker. Left: a global camera motion was experienced during frame $t$ and $t+1$, our global scan matched the object that the local correlation missed.
  Right: with our motion margin loss, the detection score of motion-blurred object is increased.}
  \label{fig:challenges}
  \vspace{-1.8em}
\end{figure}
Firstly, both the ground object motion and the aerial camera motion exist in the UAVMOT task, posing a challenge for the tracker to follow the object accurately.
Secondly, the severe motion blur~\cite{kurimo2009effect} caused by camera motion will significantly increase the difficulty of object detection.

Kalman filtering and its various variants are very commonly used motion modeling schemes~\cite{bewley2016simple,cao2023observation,zhang2022bytetrack,liu2023uncertainty}. However, the Kalman filtering scheme is a non-learning algorithm, which can only make pre-assumptions about the motion (e.g., assume linear motion), therefore, the accuracy is limited in camera motion scenes.
Several studies~\cite{shuai2021siammot,zhou2020tracking,yao2023folt} propose learning-based motion modeling, but most of them are based on local cross-correlation or local convolution and ignore global motion information.
Therefore, the lack of global motion modeling limits the tracking accuracy of these trackers in scenes with significant global camera motion.
Furthermore, previous studies have ignored the motion long-tailed distributions present in UAV-MOT datasets. Consequently, objects with significant motion, which are difficult to detect, receiving less training compared to easy-to-detect objects that exhibit minimal motion.

To address these issues, we propose MM-Tracker for fast and accurate motion modeling and motion-blur-focused object detector training.
Unlike previous work that redundantly extracts object detection features and motion features from raw input images, MM-Tracker estimates object motion from bi-temporal object detection features, which greatly reduces the computational effort of motion modeling.
In addition, MM-Tracker improves the detection performance in UAV-MOT scenes by proposing a Motion Margin loss function to impose a larger classification loss for objects with larger motions, thus effectively improves the detection of motion-blurred objects.
The main contributions of this paper are summarized as follows:
\begin{itemize}
 \item We propose the Motion Mamba module, which models object motion by local correlation of detection features and global scan of bi-directional mamba block, reached fast and accurate motion modeling.
     \item We propose the Motion Margin loss (MMLoss), which imposes a large decision boundary for large motion objects to supervise the object detector, and improves the detection performance on motion blurred objects.
     \item The MM-Tracker proposed based on the Motion Mamba module and Motion Margin loss reached state-of-the-art on two public UAV-MOT datasets, which promoted the progress of multiple object tracking in UAV scenes.
\end{itemize}

\section{Related Works}
\subsection{Motion Modeling in MOT}
Existing MOT methods typically follow the detection and association scheme~\cite{luo2021multiple,fagot2016improving,peng2020tpm,peng2020dense}.
The association criterion can be classified into two types: positional information~\cite{zhao2012tracking,forsyth2006computational,takala2007multi} and appearance information~\cite{sugimura2009using,li2013survey,yang2012multi,peng2018tracklet}. 
Association based on positional information needs to model the object motion to predict the position of the object at the next moment~\cite{bewley2016simple,shuai2021siammot,zhou2020tracking,zhang2022bytetrack,bochinski2017high}, while methods based on appearance information need to extract the appearance information of the object to compute the appearance similarity of the object between frames~\cite{wojke2017simple,zheng2016mars,wang2020towards,zhang2021fairmot}.
The small object size, blurred object appearance, and very similar appearance between ground vehicles make appearance information unreliable for object association in UAV-MOT scenes~\cite{yao2023folt}.
Therefore, this work focuses on motion modeling as association.

The most commonly used motion modeling methods in MOT are Kalman filtering (KF)~\cite{kalman1960new} and its various variants~\cite{bewley2016simple,cao2023observation,zhang2022bytetrack}. 
These methods require prior assumptions about object motion patterns and their probability distribution without any learnable parameters, which limits their performance in scenes with human-made global camera motions.
Several studies~\cite{zhou2020tracking,shuai2021siammot,yao2023folt} introduce learnable neural networks to learn the motion of objects.
These methods perform better than those based on preset rules.
However, they mainly consist of local convolution and cross-correlation, therefore limited for global motions.
In addition, these methods also have the problem of repeated feature extraction, which introduces redundant computations.

\subsection{Effective Global Information Aggregation}
Several studies~\cite{azad2019bi,wang2018deepstcl,huang2021spatial} combine sequence models (e.g. RNN~\cite{elman1990finding}, LSTM~\cite{hochreiter1997long}) with convolution to extract global image features.
However, their recursive inference scheme makes them difficult to train in parallel, resulting in slower training speed.
Their stacked sequence architectures are also prone to problems such as gradient explosion or vanishing.
By introducing a global attention mechanism, Transformer model~\cite{vaswani2017attention} effectively overcomes the problems of low training parallelism and gradient vanishing, therefore achieving success in long sequence text tasks such as language modeling~\cite{subakan2021attention}.
However, for global information modeling of images~\cite{choi2020channel,liu2021swin,liu2022video}, the computational complexity of Transformer is too large, for example $O(n^2)$, which is prone to overfitting and difficult to reach real-time tracking.
The emergence of Mamba~\cite{gu2023mamba} brings a more efficient global information extraction scheme, which can realize linear-time global attention calculation while also performing parallel training well.
Therefore, this paper adopts the bidirectional scanning Mamba structure to compute global information aggregation and thus predict long-range object motion better.

\begin{figure*}[ht]
\centering
\includegraphics[width=\textwidth]{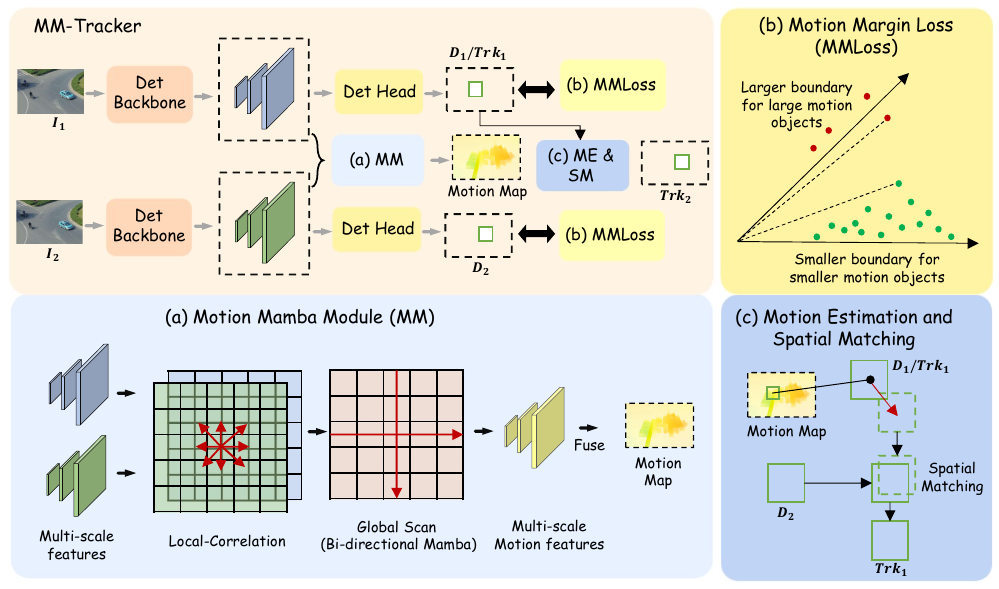}
\vspace{-2em}
\caption{Overall architecture of MM-Tracker. Multi-scale detection features are first extracted using a detection backbone (DetBackbone), which is fed into the detection head (DetHead) to output the object bounding box, score, and category. The object score is optimized using the proposed MMLoss.
The detection feature is also fed into the proposed Motion Mamba module (MM), which captures the difference between the two detection features and predicts the motion map. Afterward, the position of the object in the previous frame in the next frame is predicted based on the motion map, and the predicted position of the object is matched with the detected position in the current frame to generate a new object trajectory.}
\label{fig:net_architecture}
\vspace{-1.5em}
\end{figure*}

\subsection{Imbalance Training Loss}
The conventional imbalance training~\cite{wang2020devil} refers to the imbalance between different classes that multiple classes only account for a small proportion of the dataset, while a few classes account for a large proportion of the dataset. This imbalance results in classes with few training samples not being adequately trained, affecting the classification accuracy.
Many scholars have proposed solutions to this problem, such as
resampling~\cite{huang2016learning,li2020overcoming,ren2020balanced},
Loss reweighting~\cite{lin2017focal,li2021generalized,li2022equalized}, re-margining~\cite{cao2019learning,wang2018additive,menon2020long}.

In the UAVMOT scene, the camera's perspective change will cause large movements of objects, and this perspective change is accidental, making this situation account for a small proportion of the dataset.
However, the large motion introduces severe motion blur on objects, requiring us to focus more on those difficult-to-detect objects, which is missed in previous studies.
To this end, we propose Motion Margin loss, which imposes larger classification boundaries for objects with larger motions, thus better solving the problem of less training for large motion objects.

\section{Proposed Method}
\label{chap:proposed_method}
\subsection{Overview}
Aiming at efficient global motion modeling with a fast tracking speed, we design a Motion Mamba block with a feature reuse scheme.
Furthermore, to better detect motion-blurred objects, we propose Motion Margin loss function to supervise the classification branch of the object detector.
Fig.2 shows the overall architecture of our proposed MM-Tracker. The DetBackbone extracts detection features from each frame of the video, which are input to DetHead for object detection and Motion Mamba module for motion feature extraction.
The DetBackbone generates three scale feature maps, whose sizes are 1/8, 1/16, and 1/32 of the original image sizes. The DetHead detects objects at these three scales respectively. The classification branch of the detector is supervised by the proposed MMLoss, and the regression branch is supervised by IOU Loss and L1 Loss.
The Motion Mamba module extracts motion features from bi-temporal detection features at three scales respectively. 
The module then upsamples the features from the lowest scale step by step and fuses it with the higher scale, then generates a motion feature map with a size of 1/8 of the original image.
Finally, this feature map is supervised by L1 Loss with the ground-truth motion map.
The generation way of ground-truth motion map is illustrated in Fig.~\ref{fig:gt_generation}. 
The DetBackbone and DetHead are  YOLOX's backbone and YOLOX's head respectively.

\subsection{Motion Mamba Module}
\begin{figure}[ht]
  \centering
  \includegraphics[width=\linewidth]{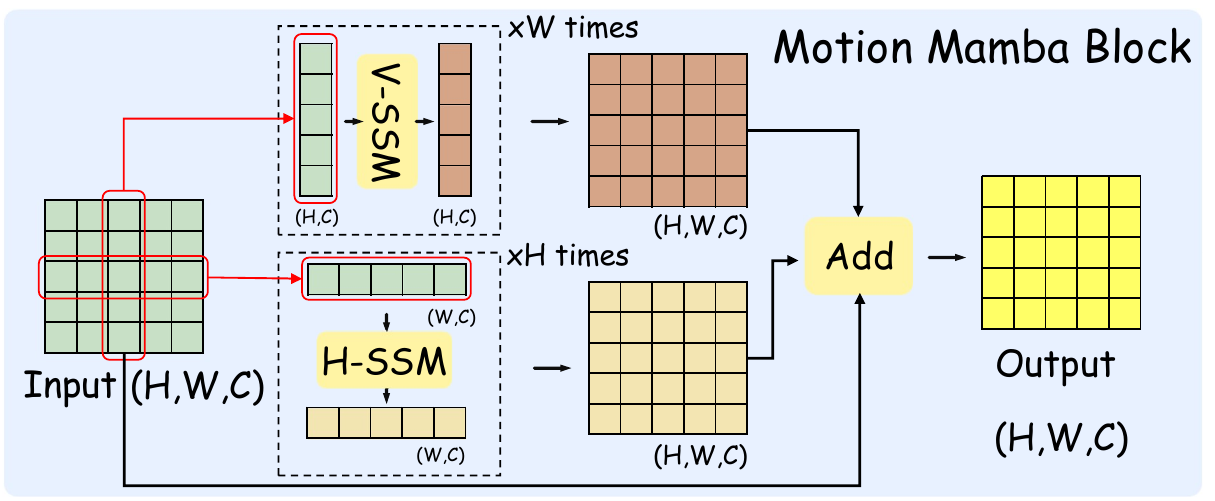}
  \vspace{-1.8em}
  \caption{Structures of our Motion Mamba block. The vertical state space model (V-SSM) and the horizontal state space model (H-SSM) are used to scan the feature map in two directions. The scanned results are added to realize global feature interaction. The short-cut connection is adopted to accelerate training.}
  \label{fig:motion_mamba}
\end{figure}

The Motion Mamba module is dedicated to efficient and lightweight motion modeling. 
Therefore, the module extracts motion features from the existing detection features of the previous and next images, which greatly reduces the computational complexity of the model.
The Motion Mamba module extracts motion features at three scales. These multi-scale features are then fused step by step from the lowest scale upwards, and finally outputs a motion feature map with a size of 1/8 of the original image.
For each scale, Motion Mamba first uses the cross-correlation of the feature maps at the previous and next moments to extract the local motion information, and then
use Motion Mamba block to extract global motion features.
As Fig.~\ref{fig:motion_mamba} shows, each Motion Mamba block consists of two branches: vertical state space model (V-SSM) and horizontal state space model (H-SSM).
These two branches conduct selective scanning~\cite{gu2023mamba} in vertical and horizontal directions respectively. 
In order to fully scan the entire feature map while reducing the computational complexity, each scan uses the vector at the pixel position as the basic unit.
Specifically, for a feature map with height, width, and channel (H, W, C) respectively, the vertical scan starts from the first row of each column, with a feature vector of size (1, 1, C) as input, and scans to the final row. The scanning of each column is performed separately, and a total of W groups of scans are performed. The horizontal SSM scans a total of H groups, follow the same scheme as the vertical scan.
The vertical-scanned feature map and the horizontal-scanned feature map are added to realize global feature interaction. The short-cut connection of the original input to the output is also adopted to accelerate training.
The top layer of feature maps outputs a feature map of size (1/8H, 1/8W, 2) through convolution, where the first channel represents the horizontal motion and the second channel represents the vertical motion.

\noindent \textbf{State Space Model}
The State Space Model~\cite{gu2023mamba} used in our Motion Mamba is calculated by an iterative computation of a sequence:
\begin{equation}
\begin{split}
        h_t = \hat{A}h_{t-1} + \hat{B}x_t, \\
        y_t=Ch_t + Dx_t,
\end{split}
\label{eq:1}
\end{equation}

with $x_t$ denote the input at time $t$, $h_t$ denote model's hidden state at time $t$, $y_t$ denote the output at time $t$, $C, D$ are model parameter matrix, $\hat{A}, \hat{B}$ are defined by:

\begin{equation}
    \hat{A} = exp^{Adt}, \hat{B} = Bdt, dt = MLP(x_t),
\label{eq:2}
\end{equation}
where $A,B$ are model parameter matrix, $dt$ is depend on our input $x_t$, MLP denote a multiple layer perceptron.
Eq.~\eqref{eq:2} reveals that the $\hat{A}$ and $\hat{B}$ matrices depend on the model input $x_t$, which reflects the selective scanning scheme of Mamba.
Scanning the entire sequence also ensures that the model can model the long-range dependencies of the feature map, which is better than only focusing on local information.

\begin{figure}[ht]
  \centering
  \includegraphics[width=\linewidth]{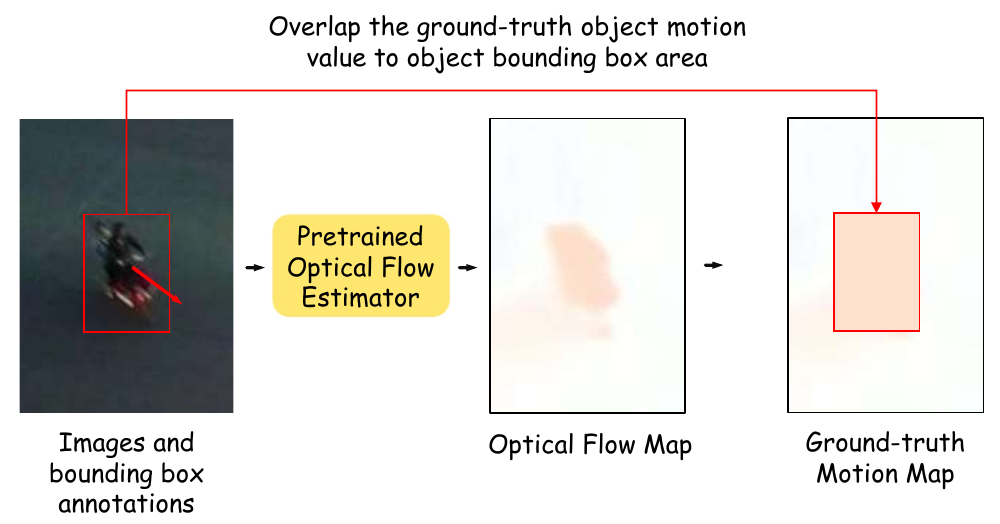}
  \vspace{-1.8em}
  \caption{Ground-truth motion map generation procedures.}
  \label{fig:gt_generation}
  \vspace{-1.em}
\end{figure}

\noindent \textbf{Motion Ground-truth Generation.}
As Fig.~\ref{fig:gt_generation} shows, we first use the pre-trained EMD-Flow~\cite{deng2023explicit} network to generate the optical flow map of each frame in the dataset.
Then we calculate the center offset of each annotated object in the dataset between its previous and next frames and overlap the offset value within the bounding box range of the object on the original optical flow map.
This map is used as ground-truth motion map to supervise the Motion Mamba module.
The introduced optical flow map helps to accelerate the convergence of the motion model.
The usage of the object motion ground-truth also makes full use of the annotation information of the original dataset.

\noindent \textbf{Motion Prediction and Tracking.}
For the object detected at time $t$, we use its center position value on the motion feature map as the motion of the object between time $t$ and $t+1$. This motion value is used to update the position of the object. The predicted object position at time $t+1$ is spatially matched with the position of the object detected by the detector at time $t+1$. This matching gives the final tracking result at time $t+1$.

\subsection{Motion Margin Loss}
The rotation of the camera view introduces a large motion of the object in the image, resulting in severe motion blur.
This motion blur will greatly increase the difficulty of object detection.
However, since there are fewer such cases in the dataset, these difficult-to-detect samples have fewer training times than easy samples, which further increases the detection difficulty.
For object tracking tasks, even a few frames that cannot be detected will cause tracking interruption, greatly affecting tracking accuracy.
To this end, we propose a Motion Margin loss function to assign different decision boundaries according to different object's motion.
We assign larger decision boundaries to objects with larger motion, thereby forcing the model to output higher scores for objects with larger motion during the learning process, so as to effectively detect these objects during inference.
Specifically, we calculate the motion margin of each object frame based on the object offset  $x$:
\begin{equation}
    D(x) = s*(\frac{1}{1+e^{-(x-5)/s}}) - M,
\end{equation}
where $M$ is defined by:
\begin{equation}
    M = s*(\frac{1}{1+e^{-(0-5)/s}})
\end{equation}

\begin{figure}[ht]
  \centering
  \includegraphics[width=\linewidth]{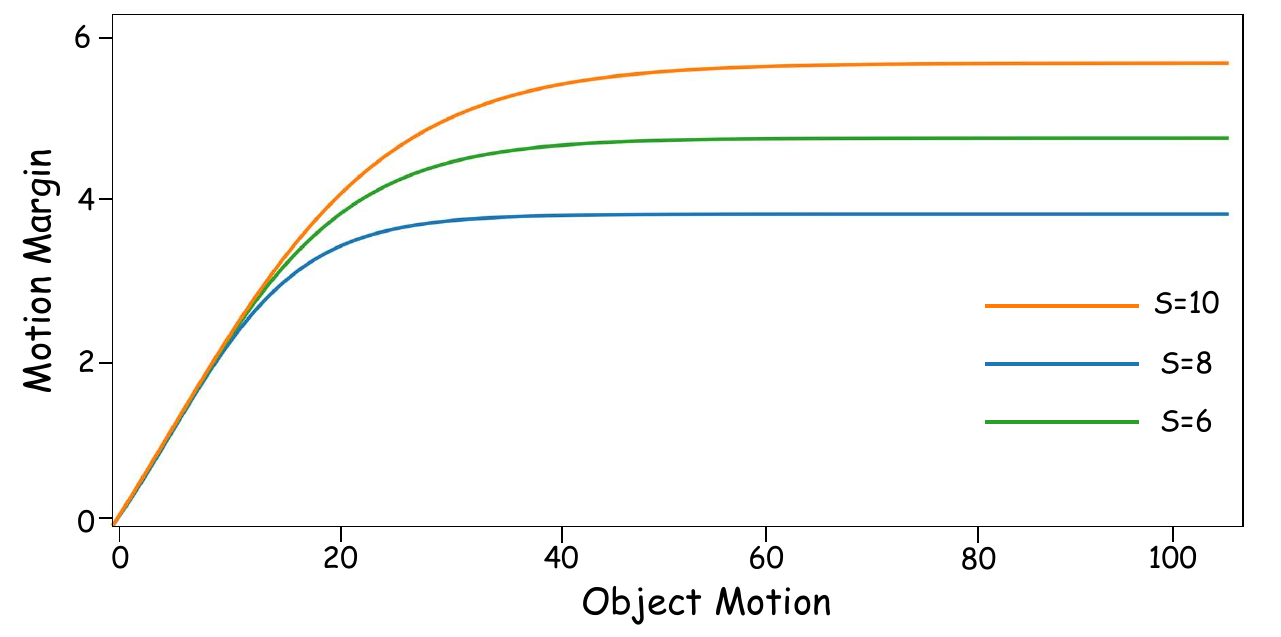}
  \vspace{-1.8em}
  \caption{Motion margin function curves of MMLoss.}
  \label{fig:offset_margin}
  \vspace{-0.5em}
\end{figure}

The design of the motion margin function follows these criteria:
(1) No margin is set when the object motion is 0.
(2) The larger the object's motion, the larger the motion margin value.
(3) The motion margin gradually converges as object motion increases.
Fig.~\ref{fig:offset_margin} shows the motion margin function curves with different s values, which indicate that the function satisfied these three criteria.

Starting from the observation that a motion of about 30 pixels can cause a large motion blur, we select s=10 as the parameter of MMLoss. Fig.~\ref{fig:offset_margin} shows that when s=10, the motion margin tends to saturation above 30.

For the classification of each object box, the motion margin loss is defined as:
\begin{equation}
\begin{split}
    {\rm MMLoss}(y_i,\hat{y_i},D_i) = -y_i*{\rm log}(\sigma(\hat{y_i}-D_i))-\\
     (1-y_i)*{\rm log}(1-\sigma(\hat{y_i})),
\end{split}
\label{equ:MMLoss}
\end{equation}
where $\hat{y_i}$ is the predicted class probability, $y_i$ is the ground-truth label. The function of subtracting $D_i$ from the output of the network classification layer is to assign different decision boundaries to different object boxes based on their motion value.

\section{Experiments}
\label{chap:experiments}
\subsection{Datasets and Metrics}
\textbf{Datasets.} 
We conduct comparative experiments on the Visdrone~\cite{zhu2020detection} and UAVDT~\cite{du2018unmanned} datasets.
These two datasets are open-sourced multi-class multi-object tracking datasets, and both are collected from the perspective of UAVs. Therefore, it is suitable to study the tracking objects with global motion and motion blur problems in these two datasets.

The Visdrone dataset consists of a training set (56 sequences), validation
set (7 sequences), test-dev set (7 sequences), and test-challenge set (6 sequences).
There are 10 categories in the Visdrone dataset: pedestrian, person, car, van, bus, truck, motor, bicycle, awning-tricycle, and tricycle. Each object within the above categories is annotated by a bounding box, category number, and unique identification number.
In experiments of Visdrone, we use the full ten categories in training while only using five categories in testing,
i.e., car, bus, truck, pedestrian, and van in evaluation, as the evaluation toolkit offered by Visdrone officials only evaluates these five categories.

The UAVDT dataset is a car-tracking dataset based on aerial view. It includes different common
scenes, such as squares, arterial streets, and toll stations.
UAVDT dataset consists of a training set (30 sequences), and a test set (20 sequences), with three categories: car, truck, and bus. In experiments of UAVDT, all three categories are evaluated using Visdrone's official evaluation toolkits.
 
\noindent\textbf{Metrics.} 
We select MOTA and IDF1 as our main evaluation metrics.

\begin{table}[ht]
\small
\centering
\setlength\tabcolsep{2pt}
\begin{tabular}{ccc|cc|cc|c}
\toprule
         &         &              & \multicolumn{2}{c|}{Visdrone}    & \multicolumn{2}{c|}{UAVDT}       &                      \\
         \midrule
B & MML  & MMM & MOTA$\uparrow$ & IDF1$\uparrow$ & MOTA$\uparrow$ & IDF1$\uparrow$ & Time(ms)$\downarrow$ \\
\midrule
$\surd$  &         &              & 39.6           & 50.4           & 47.5           & 63.1           & 15.2                 \\
$\surd$  & $\surd$ &              & 42.1           & 54.4           & 49.2           & 64.1           & 15.2                 \\
$\surd$  &         & $\surd$      & 40.9           & 54.5           & 49.7           & 66.1           & 6.9                  \\
$\surd$  & $\surd$ & $\surd$      & \textbf{44.7}  & \textbf{58.3}  & \textbf{51.4}  & \textbf{68.9}  & \textbf{6.9}        \\
\bottomrule
\end{tabular}
\vspace{-1.em}
\caption{Ablation studies on Visdrone and UAVDT test set. The $\uparrow (\downarrow)$ means that the higher (lower) result is better. The best result is marked in bold. B: Baseline model. MML: Motion Margin Loss. MMM: Motion Mamba module.}
\label{tab:ablation}
\end{table}

\begin{figure}[t]
\centering
\includegraphics[width=0.5\textwidth]{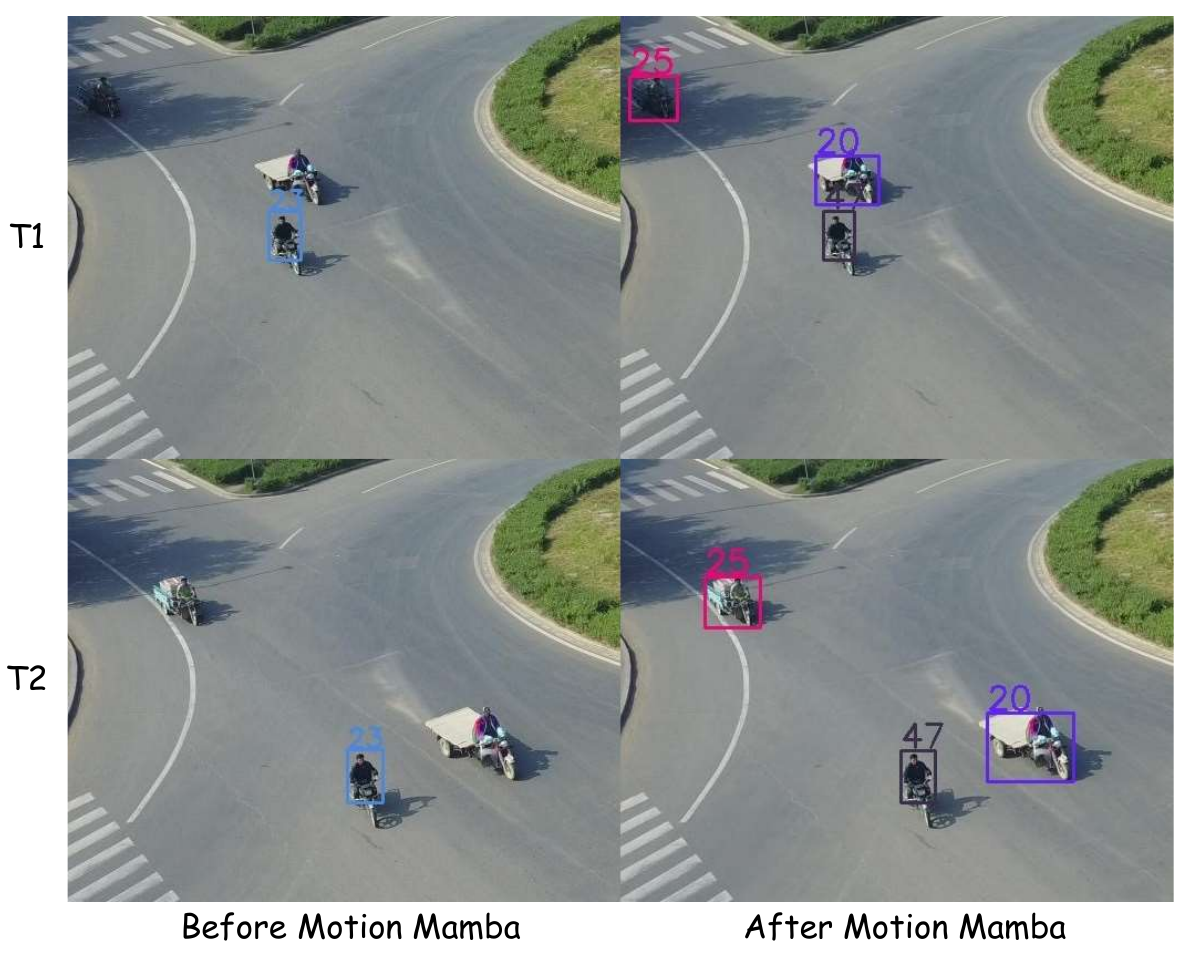}
\vspace{-2.0em}
\caption{Visualization results for large motion scenes. The model without Motion Mamba missed the fast-moving tricycles on the road, while the model with Motion Mamba tracked them with ID 20, 25, and 47.}
\label{fig:ablation_FDM}
\end{figure}

\subsection{Implementation Details}
In all experiments, we keep the same train-test split as the official split of the Visdrone and UAVDT datasets.
We use the YOLOX-S~\cite{ge2021yolox} model as the base object detector on both datasets, the input image size is $1088\times 608$.
We use the stochastic gradient descent method~\cite{bottou2010large} to optimize the detector, the learning rate is set to $0.0001$, the batch size is set to 8, each dataset is trained for 10 epochs, and the training and testing are completed on a single 2080TI graphic card.
The Motion Mamba module and the object regression branch of the detector are trained using the L1 loss function.
The object classification branch of the detector is trained using the proposed MMLoss.
We select the accuracy evaluation tool officially provided by the Visdrone dataset to complete the comparison of all metrics.
\begin{figure}[ht]
\centering
\includegraphics[width=0.5\textwidth]{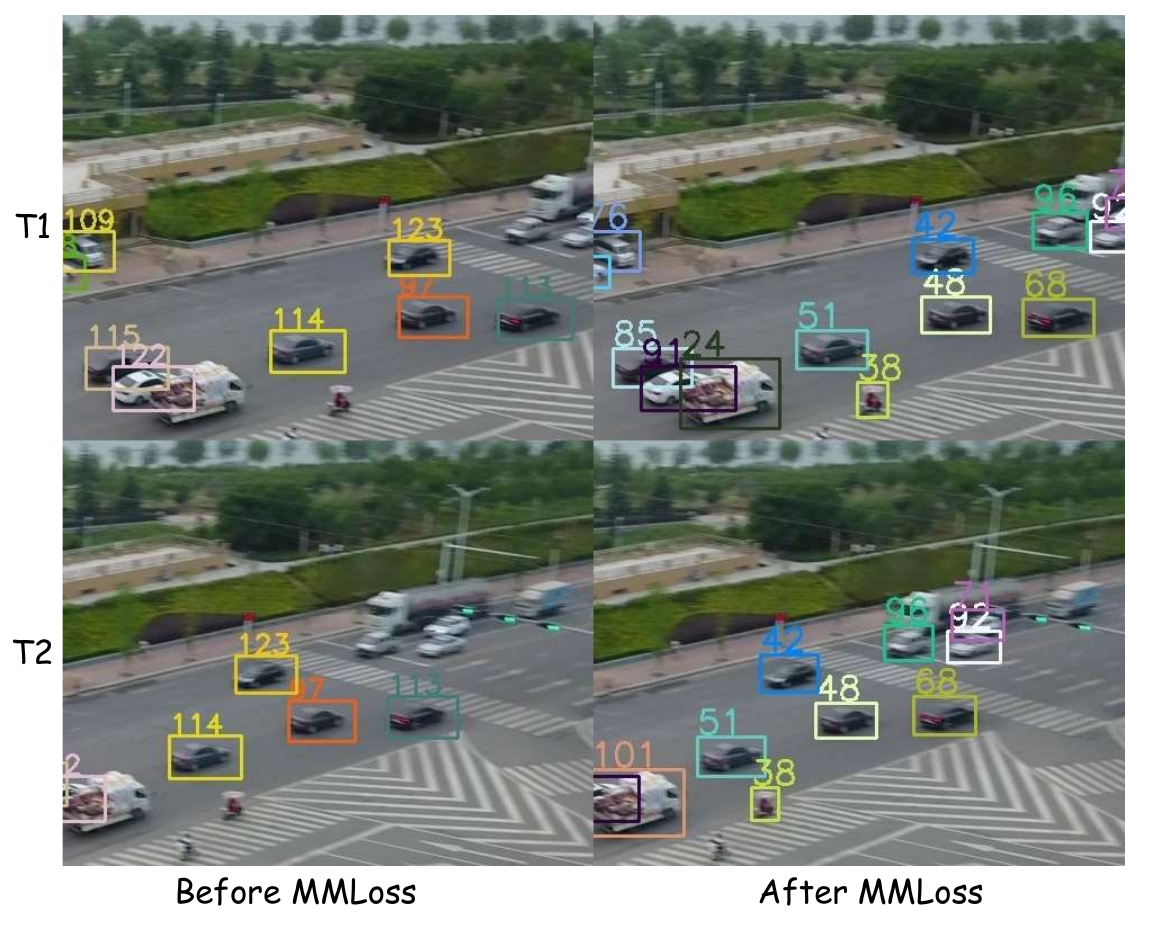}
\vspace{-2.0em}
\caption{Visualization results for blurred motion scenes. The model without MMLoss missed the fast-moving bikes on the crossroad while the model with MMLoss tracked it with ID 38.}
\label{fig:ablation_fgml}
\end{figure}

\begin{table}[htbp]
\small
\begin{tabular}{cccc}
\toprule
Motion modeling & MOTA$\uparrow$ & IDF1$\uparrow$ & Time(ms)$\downarrow$ \\
\midrule
OC-SORT       & 39.6 & 50.4 & 15.2     \\
\midrule
FastFlowNet  & 40.4 & 53.1 & 11      \\
GMA          & 40.5 & 53.5 & 102      \\
EMD-Flow     & 40.7 & 54.0 & 115      \\
Motion Mamba (Ours) & \textbf{40.9} & \textbf{54.5} & \textbf{6.9}      \\

\bottomrule
\end{tabular}
\vspace{-0.9em}
\caption{Effectiveness of Motion Mamba module. The best result is marked in bold. The $\uparrow (\downarrow) $ means that the higher (lower) result is better.}
\vspace{-1.5em}
\centering
\label{tab:ablation_FDM}
\end{table}

\subsection{Ablation study}
\noindent\textbf{Baseline model.} The baseline model we compared with is the model that uses YOLOX-S as the object detector, extended Kalman Filter as motion modeling~\cite{cao2023observation}, and two-stage spatial matching as association~\cite{zhang2022bytetrack}.

\noindent\textbf{Motion Mamba Module.}
We compare Motion Mamba with the OC-SORT tracker, and direct motion prediction using the optical flow network. As shown in Table ~\ref{tab:ablation_FDM}, the tracking accuracy obtained by Motion Mamba is better than that of OC-SORT, and is also better than directly using the optical flow network to predict motion (40.9 vs 40.7, 39.6 in MOTA, 54.5 vs 54.0, 50.4 in IDF1).
The results of EMD-Flow are close to those of Motion Mamba, but the inference time of MotionMamba is only 1/16 of that of EMDFlow (6.9 vs 115 in ms), which demonstrates the efficiency of Motion Mamba.

As Fig.~\ref{fig:ablation_FDM} shows, after using our Motion Mamba, the tracker successfully tracks the three fast-moving tricycles on the road (with ID1 20, 25, 47), while the model without Motion Mamba only tracked one tricycle (with ID 23).

We also tested the effectiveness of local and global information. 
As Table~\ref{tab:local_global} shows, 
both local correlation and global SSM improves the tracking accuracy.
Combining local correlation with two SSMs on different directions achieves the best tracking accuracy (40.9 in MOTA and 54.5 in IDF1).
\begin{table}[htb]\
\centering
\begin{tabular}{lcc}
\toprule
                        & MOTA & IDF1 \\
\midrule
Baseline                & 39.6 & 50.4 \\
Local                   & 40.2 & 52.5 \\
Local \& V-SSM          & 40.5 & 53.0 \\
Local \& H-SSM          & 40.5 & 53.5 \\
Local \& H-SSM \& V-SSM & \textbf{40.9} & \textbf{54.5} \\
\bottomrule
\end{tabular}
\vspace{-0.9em}
\caption{Effectiveness of local correlation (Local), vertical-state-space-model (V-SSM), and horizontal-state-space-model (H-SSM). The best result is marked in bold.}
\label{tab:local_global}
\vspace{-1em}
\end{table}
In summary, the qualitative and quantitative results above confirm the effectiveness of our proposed Motion Mamba in improving object tracking with large motion.


\noindent\textbf{Motion Margin Loss.}
We compare the accuracy of the proposed Motion Margin loss (MMLoss) with other imbalance training loss functions. Table ~\ref{tab:loss_comparison} shows that the MMLoss proposed in this paper performs better than the previous category long-tailed loss functions (MOTA reaches 42.1, compared to 40.3 36.4, and IDF1 reaches 54.4, compared to 51.0 and 48.3), which proves the effectiveness of our proposed MMLoss.

\begin{table}[ht]
\centering
\small
\vspace{-1em}
\begin{tabular}{ccc}
\toprule
Loss function & MOTA$\uparrow(\%)$ & IDF1$\uparrow(\%)$ \\
\midrule
Cross Entropy & 39.6 & {50.4} \\
Focal    & 36.4 & 48.3 \\
LDAM      & 40.3 & 51.0 \\
MMLoss (Ours)     & \textbf{42.1} & \textbf{54.4} \\
\bottomrule
\end{tabular}
\vspace{-0.9em}
\caption{Comparisons of different imbalance training loss functions in Visdrone test-dev dataset. The $\uparrow$ means that the higher result is better. The best result is marked in bold.}
\label{tab:loss_comparison}
\vspace{-1em}
\end{table}

\begin{figure*}[htbp]
\centering
\includegraphics[width = \textwidth]{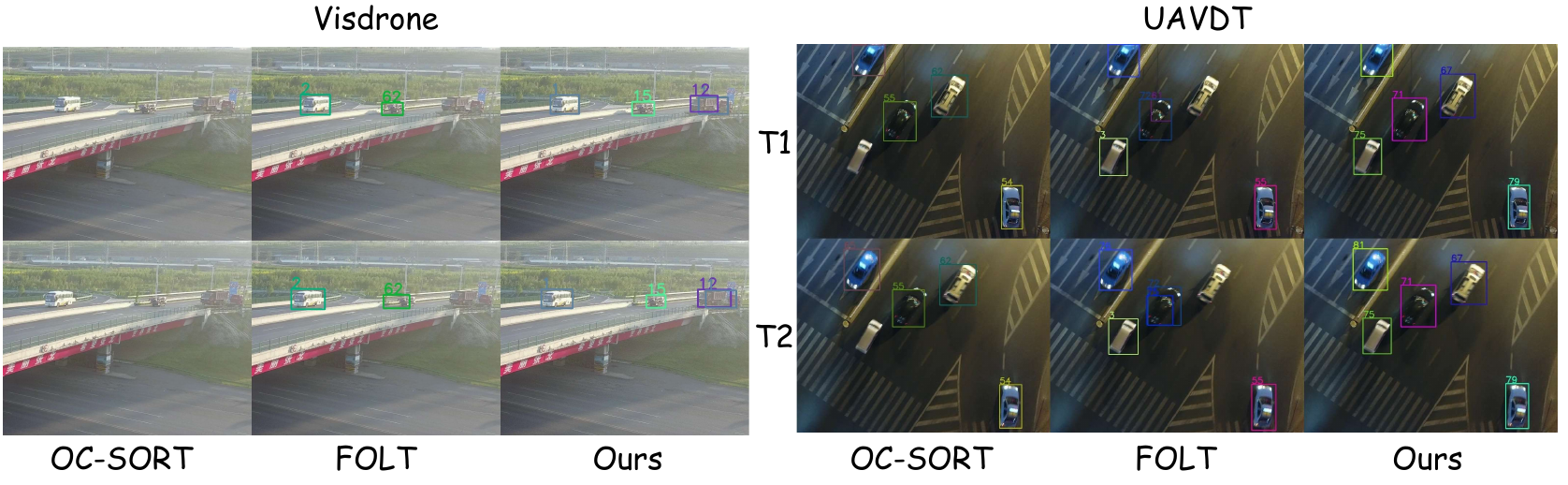}
\vspace{-1.5em}
\caption{ 
Visual comparisons on the Visdrone and UAVDT datasets. The same number denotes the same object in different frames.
}
\vspace{-1.5em}
\label{fig:comparisons_visdrone_uavdt}
\end{figure*}

Fig.~\ref{fig:ablation_fgml} visualized our ablation results.
As shown in Fig.~\ref{fig:ablation_fgml}, the model using MMLoss successfully tracked the moving bike-rider with blurred appearance (denoted as ID 38), while the baseline model failed to track it.
The qualitative and quantitative results above confirm the effectiveness of our proposed loss function.

\begin{figure}[t]
\centering
\includegraphics[width=0.5\textwidth]{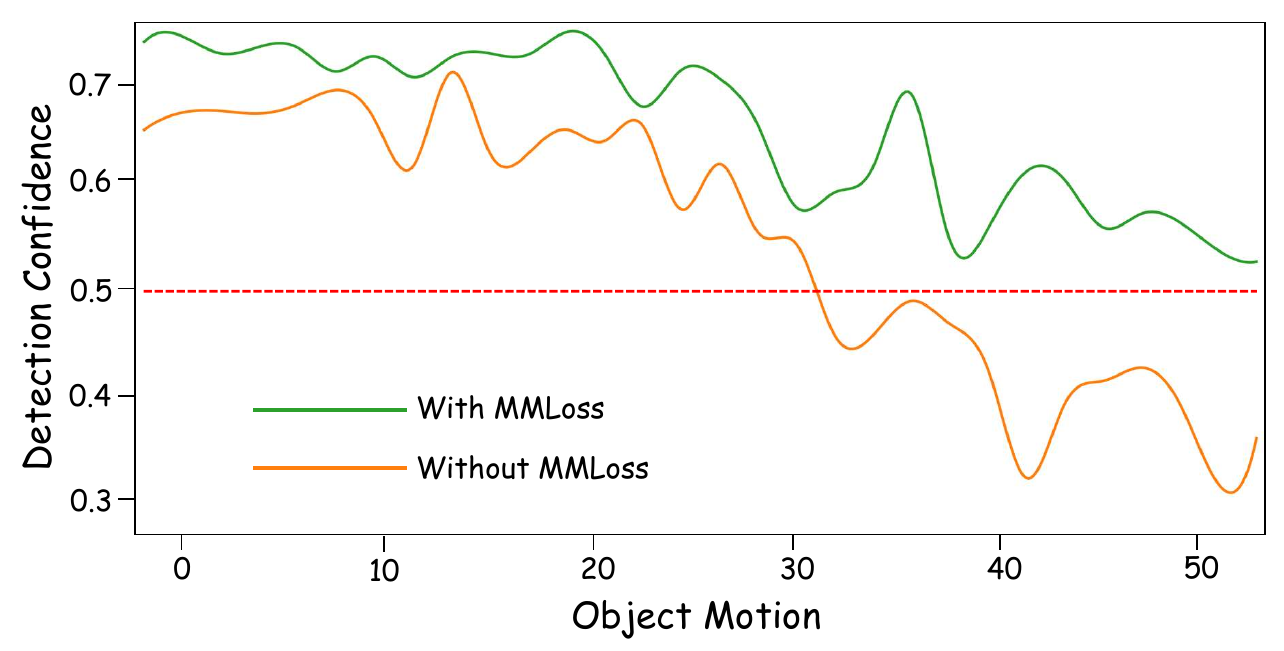}
\vspace{-1.5em}
\caption{Visualization of object's detection scores with different training loss. With MMLoss introduced, the average detection score of large motion objects is higher than 0.5 (denoted as the red dash line), which is more reliable than the model without MMLoss.
}
\label{fig:visual_margin}
\vspace{-2.0em}
\end{figure}

We further investigate the effectiveness of MMLoss to improve the detection of fast-moving objects.
As Fig.~\ref{fig:visual_margin} shows, introducing MMLoss effectively increases the average object scores of large motion objects (objects with larger relative velocity). When using MMLoss, the average score of large motion objects is higher than 0.5, which is more reliable than the model without MMLoss.

\noindent\textbf{Combination.}
We also evaluate the effectiveness of our MMLoss and Motion Mamba which are used in combination.
As Table~\ref{tab:ablation} shows, both Motion Mamba and MMLoss can effectively improve the tracking performance on the Visdrone dataset ($39.6\%$ to $40.9\%$ and $39.6\%$ to $42.1\%$ in MOTA, $50.4\%$ to $54.5\%$ and $50.4\%$ to $54.4\%$ in IDF1) and UAVDT dataset ($47.5\%$ to $49.7\%$ and $47.5\%$ to $49.2\%$ in MOTA, $63.1\%$ to $66.1\%$ and $63.1\%$ to$64.1\%$ in IDF1).
With the MMLoss and Motion Mamba commonly used, the best tracking performance is obtained (44.7\% in MOTA and 58.3\% in IDF1 on Visdrone, 51.4\% in MOTA and 68.9\% in IDF1 on UAVDT), and the motion-inference time is lower than the baseline model (15.1 to 6.9 in ms).

\begin{table}[ht]
\small
\begin{center}

\setlength\tabcolsep{1pt}

\begin{tabular}{c|cc|cc|c}
\toprule
                                                          & \multicolumn{2}{c|}{Visdrone}    & \multicolumn{2}{c|}{UAVDT} &  \\
\midrule
Method                                                    & MOTA$\uparrow$ & IDF1$\uparrow$ & MOTA$\uparrow$ & IDF1$\uparrow$ & FPS$\uparrow$ \\
\midrule


SiamMOT~\cite{shuai2021siammot}    & 31.9           & 48.3           & 39.4           & 61.4           & 11.2                 \\
FairMOT~\cite{zhang2021fairmot}    & 34.3           & 46.1           & 41.5           & 59.2           & 17.2                 \\
ByteTrack~\cite{zhang2022bytetrack}& 35.7           & 37             & 41.6           & 59.1           & 27.0                   \\
UAVMOT~\cite{liu2022multi}         & 36.1           & 51             & 46.4           & 67.3           & 12.0                   \\
OC-SORT~\cite{cao2023observation}  & 39.6           & 50.4           & 47.5           & 64.9           & 26.4                 \\
FOLT ~\cite{yao2023folt}           & 42.1           & 56.9           & 48.5           & 68.3           & 29.4                 \\
U2MOT~\cite{liu2023uncertainty} & 42.8           & 53.9           & 47.1           & 65.2
& 24.1 \\
TrackSSM~\cite{hu2024trackssm} & 41.9 & 55.3 & 48.1 & 65.4 & 24.1 \\
Ours                               & \textbf{44.7}           & \textbf{58.3}             & \textbf{51.4}           & \textbf{68.9}           & \textbf{31.1}   \\

\bottomrule
\end{tabular}
\end{center}
\vspace{-1.5em}
\caption{Comparisons of the proposed MM-Tracker with state-of-the-art methods on Visdrone and UAVDT test sets. The best result is marked in bold. The $\uparrow$($\downarrow$) means that the higher (lower) result is better. Our proposed MM-Tracker surpasses state-of-the-art in both tracking accuracy and inference speed.}
\label{tab:sota_comparisons}
\vspace{-1.5em}
\end{table}

\subsection{Comparison with state-of-the-art}
\vspace{-0.6em}
As shown in Table~\ref{tab:sota_comparisons}, the MM-Tracker proposed in this paper exceeds the current state-of-the-art (SOTA) methods in 
MOTA and IDF1.
We also visualize the comparison results of the latest best methods (OC-SORT and FOLT) and our MM-Tracker in the Visdrone dataset and the UAVDT dataset.
As the left part of Fig.~\ref{fig:comparisons_visdrone_uavdt} shows, our MM-Tracker successfully tracks the small and fast-moving bus and truck (with object ID 1, 15, and 12) in both $T1$ and $T2$ frames, while the OC-SORT failed to track these three vehicles, and the FOLT failed to track the half-occluded truck with ID 12. 
As the right part of Fig.~\ref{fig:comparisons_visdrone_uavdt} shows, the camera and the object are both in motion, introducing large and irregular motion and causing the object feature to blur.
In this difficult situation, our MM-Tracker successfully tracks all five cars in frame $T1$ 
, while the OC-SORT missed the left-bottom car and the FOLT has duplicated tracking of the same car with ID 72,61 at T1 time and 72,75 at T2 time.
The FOLT also missed the right-top car which is severely blurred due to large and irregular motion.

\section{Conclusion}
\label{chap:conclusion}

Aiming at efficient and fast global motion modeling, we propose the Motion Mamba module.
The Motion Mamba estimates a motion map based on features extracted by the object detector, which greatly reduces the computational cost of motion modeling.
The introduced bi-directional selective scan module extracts global motion features effectively, therefore outperforms previous motion modeling methods in UAVMOT scenes.
Aiming at addressing the motion imbalance training problems in current UAV-MOT datasets, we propose Motion Margin loss (MMLoss), which supervises the object detector according to the object's motion, and improves the detection performance on fast-moving objects.
Experiments show that both MMLoss and Motion Mamba can improve the accuracy of multiple object tracking in UAV view.
Comparison with state-of-the-art shows that our MM-Tracker achieves the best results on the two public UAV-MOT datasets in both tracking accuracy and inference speed.

\section*{Acknowledgement}
This work was supported in part by Natural Science Foundation of
China under contract 62171139, and in part by Zhongshan science
and technology development project under contract 2020AG016.

\vfill\pagebreak

\bibliography{aaai25}

\end{document}